\pdfoutput=1
\documentclass{article}
\usepackage{iclr2027_conference,times}
\usepackage{amsmath,amssymb,amsthm,mathtools}
\usepackage{graphicx,booktabs,tabularx,array,longtable}
\usepackage{xcolor,url,hyperref}
\usepackage{algorithm,algpseudocode}
\usepackage{microtype,placeins,needspace,etoolbox}
\hypersetup{colorlinks=true,linkcolor=blue,citecolor=blue,urlcolor=blue,pdfauthor={Jixin Tang, Ji Guo, Jun Zhao},pdftitle={MORAD: Reinforcement Learning with Multi-Objective Rewards for RNA Inverse Design}}
\newcommand{\E}{\mathbb{E}}
\newcommand{\clip}{\operatorname{clip}}
\newcommand{\sg}{\operatorname{sg}}
\newcommand{\mse}[1]{\left\lVert#1\right\rVert_{\mathrm{av}}^2}

\newcolumntype{Y}{>{\raggedright\arraybackslash}X}

\iclrfinalcopy
\makeatletter
\patchcmd{\@maketitle}{Published as a conference paper at ICLR 2027}{Preprint}{}{\PackageError{main}{Could not replace the ICLR page header}{}}
\makeatother
\usepackage{caption}
\ExplSyntaxOn
\tl_new:N \l__morad_table_tl
\tl_new:N \l__morad_spec_tl
\NewDocumentCommand{\eqtable}{s O{} m m}
  {
    \file_get:nnN {#3.tex} {} \l__morad_table_tl
    \regex_replace_all:nnN { tabular\* } { moradeq } \l__morad_table_tl
    \IfBooleanT {#1}
      {
        \regex_replace_once:nnN { RMSD\x20\( } { RMSD \c{\\} ( } \l__morad_table_tl
        \regex_replace_once:nnN { lDDT\( } { lDDT \c{\\} ( } \l__morad_table_tl
        \regex_replace_once:nnN { Coarse\x20clash } { Coarse \c{\\} clash } \l__morad_table_tl
      }
    \tl_if_blank:nTF {#2}
      { \tl_set:Nn \l__morad_spec_tl { @{} l *{#4}{c} @{} } }
      { \tl_set:Nn \l__morad_spec_tl { @{} l *{#4}{w{c}{#2}} @{} } }
    \tl_use:N \l__morad_table_tl
  }

\ExplSyntaxOff

\title{MORAD: Reinforcement Learning with Multi-Objective Rewards for RNA Inverse Design}
\author{Jixin Tang\thanks{Equal contribution.} \quad Ji Guo\footnotemark[1] \quad Jun Zhao\thanks{Corresponding author.} \\
Fudan University}
\begin{document}
\renewcommand{\bottomfraction}{0.5}
\maketitle
{\renewcommand{\thefootnote}{}\footnotetext{Project page: \url{https://gabrile166.github.io/MORAD/}}}
\begin{abstract}
We introduce MORAD, an online reinforcement learning framework for backbone-conditioned RNA diffusion models that trains a single shared policy across diverse target backbones. Structure-conditioned RNA inverse design seeks nucleotide sequences compatible with a target three-dimensional backbone, yet existing reinforcement learning approaches either optimize a separate policy for each target backbone or rely on offline preference optimization, leaving online post-training of a reusable backbone-conditioned model largely unexplored. MORAD combines backbone-conditioned group sampling, group-relative reward normalization, and a forward-process diffusion objective. Beyond existing RL rewards that primarily target tertiary structural quality, MORAD jointly optimizes six measurements spanning tertiary geometry, base pairing, ensemble behavior, and sequence composition. To aggregate these heterogeneous objectives despite differences in scale and optimization direction, MORAD maps each measurement to a bounded desirability score and combines them with a weighted geometric mean. Compared with a tertiary-structure-only reward, this multi-objective formulation improves 10 of the 11 evaluation metrics that have a preferred direction. Applied to pretrained RIDE, MORAD improves GDT-TS by 16.0\% (from 0.3300 to 0.3827) and pairing MCC by 19.4\% (from 0.6110 to 0.7295), while reducing normalized ensemble defect by 27.5\% (from 0.3906 to 0.2830). On the test set, MORAD outperforms RiboDiffusion, gRNAde, and RDesign across tertiary structural accuracy, secondary-structure pairing, and thermodynamic quality, establishing a new state of the art in structure-conditioned RNA inverse design.
\end{abstract}

\section{Introduction}
RNA inverse design generates nucleotide sequences that are compatible with a specified structure. Three-dimensional backbone conditioning makes local geometry and long-range spatial relationships available to a sequence generator. This formulation enables the construction of candidate RNAs for subsequent folding and functional evaluation. It also defines a one-to-many learning problem: a target backbone can admit multiple compatible sequences. Recent structure-conditioned methods therefore model a conditional sequence distribution and evaluate both structural compatibility and sequence recovery \citep{joshi2025grnade,huang2024ribodiffusion,tan2024rdesign}.

Geometric learning and diffusion modeling provide complementary ways to represent this distribution. gRNAde encodes RNA backbones with geometric graph neural networks and supports multiple conformations \citep{joshi2025grnade}. RDesign exploits hierarchical representation learning to use limited structural data efficiently \citep{tan2024rdesign}. RiboDiffusion combines structural conditioning with generative denoising, while RIDE provides a structure-conditioned diffusion backbone within the RIDER framework \citep{huang2024ribodiffusion,hu2026rider}. These models establish a strong starting distribution for design. Post-training can then further improve these models using properties measured on their newly generated sequences.

Reinforcement learning (RL) makes this feedback loop explicit. A design policy proposes candidates, a computational evaluator scores them, and the policy learns from the resulting rewards. LEARNA and Meta-LEARNA established this approach for secondary-structure-conditioned RNA design \citep{runge2019learna}. More recent RNA methods investigate preference optimization, RL-guided diffusion, and step-wise latent optimization \citep{sun2025ribopo,hu2026rider,si2026sold}. Protein design offers parallel evidence: EvoPlay combines self-play with sequence search, RL-DIF tunes structure-conditioned categorical diffusion, and ProteinZero updates inverse-folding models with online structural and stability feedback \citep{wang2023evoplay,ektefaie2024rldif,wang2025proteinzero}. Together, these developments motivate learning a reusable design policy from feedback collected across many RNA backbones.

Two coupled choices determine how such a policy learns. The first is the training interface. RNA feedback can involve a learned folding predictor, discrete base-pair comparisons, and thermodynamic software. An effective update rule must consume their scalar outputs while accommodating a multistep diffusion generator. Forward-process diffusion RL provides such an update rule through reward-weighted denoising objectives \citep{zheng2026diffusionnft}. The second choice is the reward itself. Tertiary alignment, pairing compatibility, and ensemble behavior describe different aspects of a design, with different units and scales. Their aggregation determines which candidate-level improvements the policy prioritizes. A structured reward should preserve these distinctions while producing an effective within-target learning signal.

MORAD integrates these choices into a shared-policy RNA post-training workflow. For each target, an exponential-moving-average policy samples a group of sequences. Six measurements are mapped to bounded desirability scores, which a weighted geometric mean combines into a scalar reward. Group-relative normalization converts that reward into candidate weights for a DiffusionNFT-style update \citep{zheng2026diffusionnft,shao2024deepseekmath}. Training repeats across target backbones, and the resulting policy generates sequences directly from a new backbone at inference.

Our contributions are:
\begin{enumerate}
\item \textbf{A reusable RNA online RL framework.} We formulate a shared-policy training procedure that connects structure-conditioned diffusion sampling, group-relative feedback, and forward-process optimization through modular interfaces.
\item \textbf{A coordinated multi-objective reward.} We specify six RNA design measurements, smooth desirability mappings, and weighted geometric aggregation, together with the data construction needed to support online scoring.
\item \textbf{An improved shared RNA design model.} We obtain MORAD by post-training RIDE and evaluate the same checkpoint across tertiary structure, secondary structure, and thermodynamics. On 153 evaluation targets, it reaches 0.3827 GDT-TS and 0.3400 TM-score. Held-out evaluations at 40 points during training track this improvement.
\end{enumerate}

\section{Related work}
\paragraph{Conditioned RNA design.}
Secondary-structure-conditioned design uses pairing constraints to guide sequence search or learning. LEARNA trains a design agent, and Meta-LEARNA transfers experience across target structures \citep{runge2019learna}. Three-dimensional conditioning adds backbone geometry: gRNAde develops geometric, multi-state sequence modeling; RDesign learns hierarchical structural representations; and RiboDiffusion generates sequences through conditional denoising \citep{joshi2025grnade,tan2024rdesign,huang2024ribodiffusion}. RIDER combines its RIDE backbone with RL-guided optimization, optimizing and reporting one target backbone at a time, so the resulting weights are target-specific rather than reusable; SOLD studies step-wise optimization in a latent diffusion model \citep{hu2026rider,si2026sold}. RNA3DB groups RNA structures into structurally dissimilar components, providing a basis for assessing transfer across RNA structures \citep{szikszai2024rna3db}.

\paragraph{Reward-based learning for biomolecular sequences.}
Protein RL spans mutation search, language-model adaptation, and structure-conditioned generation. EvoPlay treats residue mutations as actions and couples a policy-value network with Monte Carlo tree search \citep{wang2023evoplay}. KnowRLM incorporates biochemical knowledge into reinforced language models for directed evolution \citep{wang2024knowrlm}. RL-DIF optimizes a pretrained categorical diffusion model using structural consistency, with foldable diversity as an explicit evaluation goal \citep{ektefaie2024rldif}. ProteinZero uses online structural and stability proxies together with reference-policy and diversity regularization \citep{wang2025proteinzero}. DRAKES develops reward optimization for discrete diffusion with applications to DNA and protein design \citep{wang2025drakes}. For RNA, RiboPO constructs structure- and stability-aware preferences, complementing the online optimization routes explored by RIDER and SOLD \citep{sun2025ribopo,hu2026rider,si2026sold}.

Several methods train generative models from reward feedback. DDPO views reverse diffusion as a sequential decision process \citep{black2024ddpo}. DRAKES propagates reward information through relaxed discrete sampling, while DiffusionNFT learns through the forward process using positive and negative denoising branches \citep{wang2025drakes,zheng2026diffusionnft}. GRPO provides a group-relative baseline for reward normalization \citep{shao2024deepseekmath}. These mechanisms address distinct parts of the learning loop: trajectory optimization, reward propagation, and candidate-level credit assignment.

\paragraph{Multi-objective measurement and evaluation.}
Desirability functions map responses with different units onto a common scale; geometric aggregation is a classical construction for coordinating multiple responses \citep{derringer1980desirability}. RNA evaluation combines learned structural predictors such as RhoFold+, secondary-structure predictors such as EternaFold, and energy-based calculations from ViennaRNA \citep{shen2024rhofold,waymentsteele2022eternafold,lorenz2011vienna}. US-align supports structural comparisons of nucleic acids \citep{zhang2022usalign}. RiboPO organizes evaluation across sequence, secondary structure, tertiary structure, and thermodynamics \citep{sun2025ribopo}. Uncertainty estimates also complement aggregate RL scores \citep{agarwal2021precipice}.

\section{Method}
\subsection{Online training algorithm and optimization objective}
\begin{figure}[t]
\centering\includegraphics[width=\linewidth]{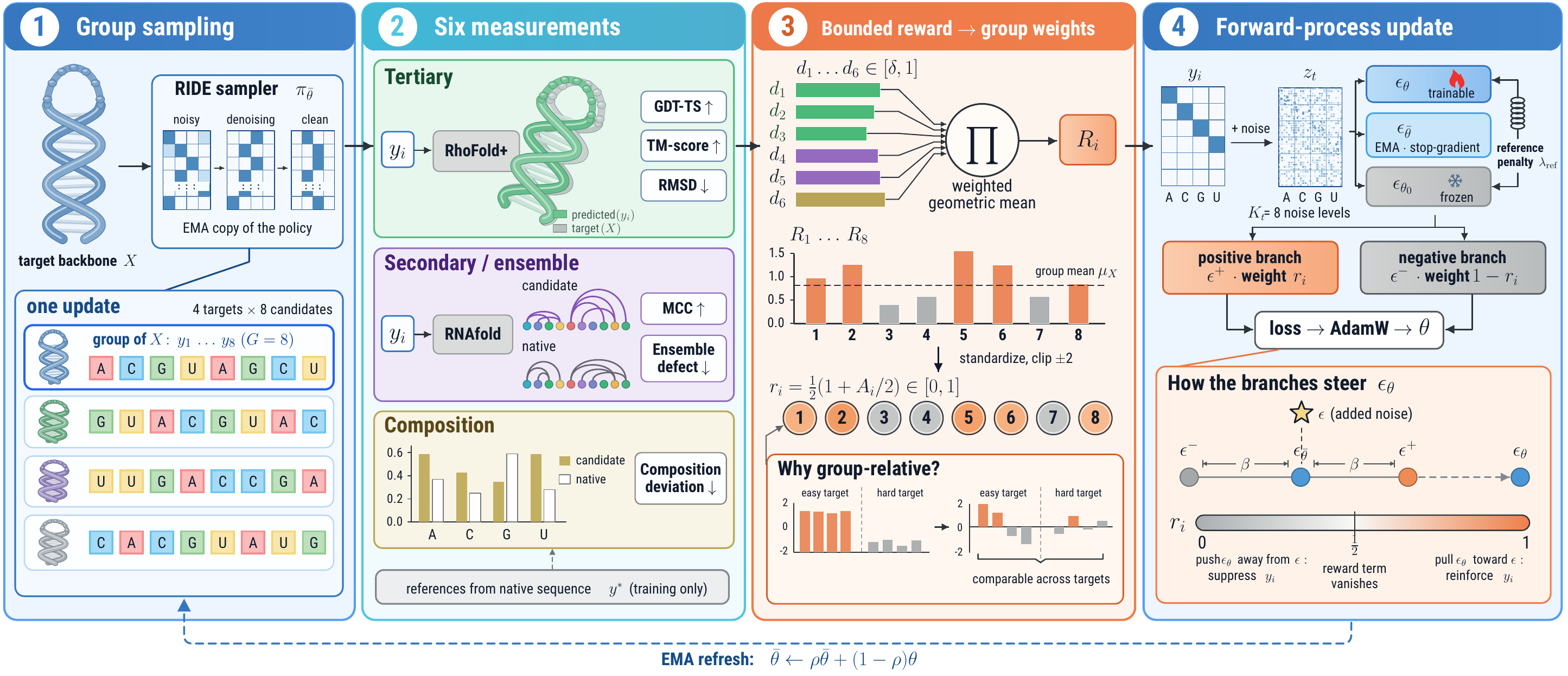}
\caption{\textbf{MORAD training loop.} Backbone-conditioned sampling produces candidate sequences. Three evaluation channels provide the six-component reward and group-relative weights. Forward-noised candidates train the shared policy under a frozen-reference constraint; EMA feedback refreshes the sampler. The update follows the positive/negative objective in Eq.~\eqref{eq:loss}; the lower feedback path denotes EMA refresh.}
\label{fig:method}
\end{figure}

Let $X$ denote an RNA backbone with $L$ valid nucleotide positions and let $y\in\{A,C,G,U\}^{L}$ denote a designed sequence. A pretrained conditional diffusion model initializes three parameter sets: the trainable policy $\theta$, the moving-average sampling policy $\bar\theta$, and the frozen reference $\theta_0$. The online objective is to increase $\E_{X\sim\mathcal D_{\rm train},\,y\sim\pi_\theta(\cdot\mid X)}[R(X,y)]$ while keeping a reference-based constraint on denoising predictions. Figure~\ref{fig:method} summarizes the training loop.

\paragraph{Target-relative feedback.}
For each sampled target, $\pi_{\bar\theta}$ generates $G$ candidates. The scorer returns valid rewards $\{R_i\}_{i\in\mathcal V_X}$. We use population statistics over the valid group and define
\begin{equation}
 \mu_X=\frac{1}{|\mathcal V_X|}\sum_{i\in\mathcal V_X}R_i,\quad
 s_X^2=\frac{1}{|\mathcal V_X|}\sum_{i\in\mathcal V_X}(R_i-\mu_X)^2,
 \quad A_i=\clip\!\left(\frac{R_i-\mu_X}{s_X+\varepsilon_A},-z_c,z_c\right).
 \label{eq:adv}
\end{equation}
The candidate weight is $r_i=\tfrac12(1+A_i/z_c)$, with $z_c=2$ and $\varepsilon_A=10^{-8}$. Group normalization compares sequences under the same structural condition and places feedback on a common scale across targets. Groups require at least two valid rewards and positive reward variance to contribute a policy gradient. Candidates are drawn at sampling temperature $\tau=1$; a group returning fewer than two valid rewards or fewer than two distinct sequences is resampled once with a new seed and the temperature raised by $0.1$, and it is dropped if the second attempt also fails. Rewards, group statistics, and $r_i$ are treated as constants during optimization.

\paragraph{Forward-process policy update.}
The one-hot sequence $x_0\in\{0,1\}^{L\times4}$ is corrupted as $z_t=\alpha_t x_0+\sigma_t\epsilon$, where $\epsilon\sim\mathcal N(0,I)$ and $\alpha_t^2+\sigma_t^2=1$ \citep{ho2020ddpm}. We sample $K_t$ stratified log-signal-to-noise levels per candidate. Let $\epsilon_\theta$, $\epsilon_{\bar\theta}$, and $\epsilon_{\theta_0}$ denote the corresponding conditional noise predictions at $(z_t,t,X)$. DiffusionNFT is formulated with flow-matching velocity prediction \citep{zheng2026diffusionnft}; we transfer its positive/negative construction to the $\epsilon$-prediction space of RIDE and form
\begin{equation}
 \epsilon^+=(1-\beta)\epsilon_{\bar\theta}+\beta\epsilon_\theta,
 \qquad
 \epsilon^-=(1+\beta)\epsilon_{\bar\theta}-\beta\epsilon_\theta,
 \label{eq:branches}
\end{equation}
with stop-gradient applied to $\epsilon_{\bar\theta}$. The scale $\beta\in(0,1]$ sets how far the two branches are displaced from the sampling policy. The objective is
\begin{equation}
 \mathcal L(\theta)=\E\!\left[
 \frac{r_i\mse{\epsilon^+-\epsilon}+(1-r_i)\mse{\epsilon^--\epsilon}}{\beta}
 +\lambda_{\rm ref}\mse{\epsilon_\theta-\sg(\epsilon_{\theta_0})}
 \right].
 \label{eq:loss}
\end{equation}
Here $\|\cdot\|_{\rm av}^2$ averages squared residuals across sequence positions and channels; the expectation averages valid candidates, targets, sampled noise, and time levels. We use uniform weighting in noise-prediction space. The reference term penalizes changes to the pretrained denoiser, and the positive/negative branches turn relative reward into an update direction. Because of the $1/\beta$ factor, the cross term that carries the reward does not depend on $\beta$, so $\beta$ scales only the quadratic displacement penalty; we use $\beta=0.1$ and $\lambda_{\rm ref}=0.02$ (Appendix~\ref{app:settings}). Appendix~\ref{app:derivation} derives this relation and the equivalent clean-sequence weighting.

We use AdamW \citep{loshchilov2019adamw}. After each optimizer step, $\bar\theta\leftarrow\rho_u\bar\theta+(1-\rho_u)\theta$, where $\rho_u=\min(0.999,(1+u)/(10+u))$ at update $u$. The frozen reference remains fixed. Algorithm~\ref{alg:train} lists each step of the training loop. It uses $\ell_{\theta,X,i,k}$ for the integrand of Eq.~\eqref{eq:loss}, $p_{\rm strat}$ for stratified time sampling, and $\operatorname{clip}_{1}$ for unit-norm gradient clipping. GroupStats returns the valid indices, reward mean, and population standard deviation. The procedure requires only a conditional sampler, a noise-prediction adapter, and scalar evaluators. The experiments instantiate these components with RIDE; the trained policy shares parameters across all target backbones.

\begin{figure}[t]
\centering\includegraphics[width=\linewidth]{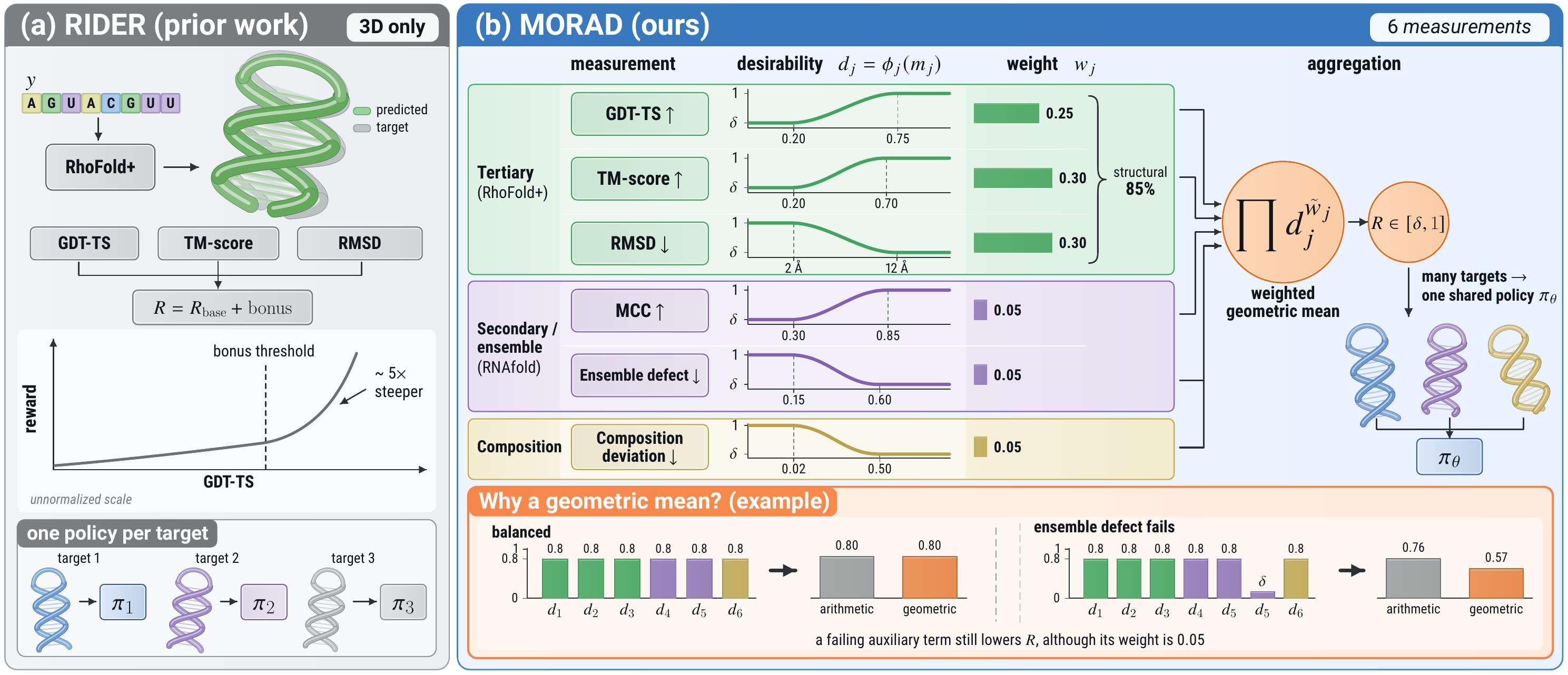}
\caption{\textbf{RIDER and MORAD reward construction.} RIDER scores tertiary structural consistency \citep{hu2026rider}. MORAD combines three tertiary measurements, pairing agreement, normalized ensemble defect (ED), and composition deviation. Each desirability score lies in $[\delta,1]$; $\phi_j$ denotes the map in Eq.~\eqref{eq:desirability}, and $\widetilde w_j$ is the normalized weight in Eq.~\eqref{eq:reward}. The panels compare reward definitions.}
\label{fig:reward}
\end{figure}

\begin{algorithm}[!t]
\caption{MORAD online post-training}\label{alg:train}
\small
\begin{algorithmic}[1]
\Require Training data $\mathcal D_{\rm train}$; pretrained parameters $\theta_0$; group size $G$; sampling temperature $\tau$;
\Statex \hspace{\algorithmicindent} noise draws $K_t$; updates $U$; loss coefficients $\beta,\lambda_{\rm ref}$
\State \textbf{Initialize policies:} $\theta\gets\theta_0,\ \bar\theta\gets\theta_0$; keep $\theta_0$ frozen.
\For{$u=1,\ldots,U$}
 \State \textbf{Sample targets:} $\mathcal B_u\sim\mathcal D_{\rm train}$.
 \State \textbf{Generate candidates:} $y_{X,1:G}\sim\pi_{\bar\theta}(\cdot\mid X)$ at temperature $\tau$ for each $X\in\mathcal B_u$.
 \State \textbf{Score designs:} $R_{X,i}\gets R(X,y_{X,i})$ using Eq.~\eqref{eq:reward}.
 \State \textbf{Resample degenerate groups once:} repeat the two steps above at temperature $\tau+0.1$ for any $X$
 \Statex \hspace{\algorithmicindent} with fewer than two valid rewards or fewer than two distinct sequences.
 \State \textbf{Summarize groups:} $(\mathcal V_X,\mu_X,s_X)\gets\operatorname{GroupStats}(R_{X,1:G})$.
 \State \textbf{Retain informative groups:} $\mathcal B_u^+\gets\{X\in\mathcal B_u:|\mathcal V_X|\ge2,\ s_X>0\}$.
 \ForAll{valid candidates $X\in\mathcal B_u^+,\ i\in\mathcal V_X$}
  \State \textbf{Normalize feedback:} compute $A_{X,i}$ by Eq.~\eqref{eq:adv}.
  \State \textbf{Detach weights:} $r_{X,i}\gets\sg\!\left(\tfrac12(1+A_{X,i}/z_c)\right)$.
  \State \textbf{Draw noise and times:} $(t_{ik},\epsilon_{ik})\sim p_{\rm strat}(t)\,\mathcal N(0,I)$, $k=1{:}K_t$.
  \State \textbf{Perturb sequences:} $z_{ik}\gets\alpha_{t_{ik}}\operatorname{onehot}(y_{X,i})+\sigma_{t_{ik}}\epsilon_{ik}$.
 \EndFor
 \State \textbf{Aggregate the two-branch loss} from Eq.~\eqref{eq:loss} over valid candidates:
 \Statex \hspace{\algorithmicindent} $\displaystyle\mathcal L_u\gets\frac{1}{|\mathcal B_u|}\sum_{X\in\mathcal B_u^+}\frac{1}{|\mathcal V_X|K_t}\sum_{i\in\mathcal V_X}\sum_{k=1}^{K_t}\ell_{\theta,X,i,k}$.
 \State \textbf{Update the policy:} $\theta\gets\operatorname{AdamW}(\theta,\operatorname{clip}_{1}(\nabla_\theta\mathcal L_u))$.
 \State \textbf{Refresh the EMA sampler:} $\bar\theta\gets\rho_u\bar\theta+(1-\rho_u)\theta$.
\EndFor
\State \Return shared design policy $\pi_\theta(\cdot\mid X)$.
\end{algorithmic}
\end{algorithm}

\subsection{Reward construction}
The reward combines six quantities with explicit roles. GDT-TS, TM-score, and RMSD assess the RhoFold+-predicted structure against the target backbone \citep{shen2024rhofold}. Secondary-structure MCC measures agreement between the RNAfold structures of a candidate and its native reference \citep{lorenz2011vienna,matthews1975mcc}. Normalized ensemble defect measures expected nucleotide-level disagreement with that reference pairing under the candidate's thermodynamic ensemble \citep{zadeh2011ensemble}. A composition-deviation term compares candidate and native A/C/G/U frequencies. Appendix~\ref{app:metrics} gives the measurement formulas and evaluation-tool configuration.

For metric $m_j$, anchors $a_j$ and $b_j$ identify low and high desirability. We apply a clipped raised-cosine map,
\begin{equation}
 u_j=\clip\!\left(\frac{m_j-a_j}{b_j-a_j},0,1\right),
 \qquad d_j=\max\!\left(\delta,\frac{1-\cos(\pi u_j)}{2}\right),
 \quad \delta=10^{-3}.
 \label{eq:desirability}
\end{equation}
Lower-is-better measurements use $b_j<a_j$. The anchors align units and specify the region where changes in a measurement affect the reward. Saturation limits the influence of extreme values, and the floor keeps logarithms finite. The scalar reward is
\begin{equation}
 R(X,y)=\exp\!\left(\sum_{j\in\mathcal J(X,y)}\widetilde w_j\log d_j\right),
 \qquad \widetilde w_j=\frac{w_j}{\sum_{k\in\mathcal J(X,y)}w_k},
 \label{eq:reward}
\end{equation}
where $\mathcal J$ contains available measurements. The weights for GDT-TS, TM-score, RMSD, MCC, ensemble defect, and composition are $(0.25,0.30,0.30,0.05,0.05,0.05)$. Valid scoring requires the three structural terms; an unavailable auxiliary measurement is dropped from $\mathcal J$, and Eq.~\eqref{eq:reward} renormalizes over the remaining active weights.

Geometric aggregation favors candidates that score reasonably on every measurement over candidates that excel on only one \citep{derringer1980desirability}. Above the desirability floor, $\partial\log R/\partial d_j=\widetilde w_j/d_j$, so at a fixed weight the reward responds most strongly to the lowest scores. This property motivates the geometric form; it reaches the optimizer through the candidate weights $r_i$. Structural terms carry 85\% of the nominal weight, while the three auxiliary terms still directly affect how candidates are ranked. With Eq.~\eqref{eq:adv}, normalization has two stages: desirability scores align the units and directions of the six measurements, and group-relative standardization removes difficulty differences between targets. Appendix~\ref{app:reward} lists all anchors and the structural-only control reward.

\begin{table}[t]
\caption{\textbf{Tertiary-structure evaluation.} First sampled candidate (candidate 0) for each of the 153 evaluation targets; C1$\prime$ TM has 126 valid targets. MORAD is the checkpoint at update 390. Alignment scores and additional diagnostics are shown together; RMSD is in \AA. Bold marks the best mean. Metric definitions are in Appendix~\ref{app:structdiagnostics}.}
\label{tab:tertiary}\centering\footnotesize\setlength{\tabcolsep}{2.9pt}\eqtable*{table_tertiary}{9}
\end{table}

\subsection{Data construction}
Each record stores a backbone $X$, its native sequence $y^\star$, and a coordinate-validity mask. The backbone conditions generation, and the native sequence provides pairing and composition references during training; at inference the policy conditions on the target backbone alone. The training set contains 527 targets and the evaluation set 153. The evaluation set is built separately from the training set and comprises 71 RNA3DB entries, drawn from RNA3DB components that contain no training target, and 82 independently curated RCSB entries spanning 15--186 nucleotides \citep{szikszai2024rna3db,berman2000pdb}. No evaluation sequence exactly matches a training sequence after normalization; each evaluation target's maximum sequence identity to the training set has a median of 0.62, and Appendix~\ref{app:data} reports its full distribution. All 153 evaluation targets are used. Appendix~\ref{app:data} details preprocessing and screening.
\FloatBarrier

\section{Experiments}
\subsection{Experimental setup}
We initialize from pretrained RIDE and train for 390 updates over three epochs on four GPUs, using four target groups of eight candidates per update. The reported model is the final checkpoint, update 390; the held-out evaluations taken every ten updates document training and are not used to select a checkpoint. The 10.2M-parameter model uses an initial learning rate of $2\times10^{-4}$ with cosine decay to 0.3 times the initial rate. RhoFold+ scores tertiary structure \citep{shen2024rhofold}; Appendix~\ref{app:settings} gives the configuration.

Tables~\ref{tab:tertiary} and \ref{tab:secondary} compare MORAD with RiboDiffusion, gRNAde ($T=0.1$), RDesign, and pretrained RIDE using the first sampled candidate (candidate 0) for each of the 153 evaluation targets \citep{huang2024ribodiffusion,joshi2025grnade,tan2024rdesign,hu2026rider}; the index is fixed for every model, with no best-of-eight selection. C1$\prime$ TM has 126 valid targets; pairing counts vary by metric (Appendix~\ref{app:paircounts}). Diversity uses all eight candidates per target (Appendix~\ref{app:seqdiv}); Table~\ref{tab:thermo} reports a separate thermodynamic evaluation that also includes gRNAde ($T=0.8$). The checkpoint comparison (Appendix~\ref{app:checkpointdetails}) and the training curves (Figure~\ref{fig:training}) come from separate evaluation runs.

\subsection{Comparison with existing RNA design models}
Across Tables~\ref{tab:tertiary}--\ref{tab:thermo}, MORAD attains the best mean on 14 of the 16 metrics with a preferred direction; the two exceptions, recovery and pLDDT, are led by RiboDiffusion.

\paragraph{Tertiary structure.}
MORAD improves GDT-TS by 16.0\% over RIDE (from 0.3300 to 0.3827) and TM-score by 19.7\% (from 0.2841 to 0.3400), while reducing RMSD from 9.7199 to 7.9196~\AA{} and raising C1$\prime$ TM from 0.3538 to 0.3855 (Table~\ref{tab:tertiary}). It also leads success rate, C4$\prime$ lDDT and coarse clash, while RiboDiffusion leads recovery and pLDDT. These metrics capture different properties: structural agreement, predictor confidence, and sequence recovery.

\begin{table}[t]
\caption{\textbf{Candidate-0 secondary-structure agreement.} EternaFold predictions are compared with base pairs extracted from the native 3D structures \citep{waymentsteele2022eternafold}. Bold marks the best mean. Valid counts for each metric and model are in Appendix~\ref{app:paircounts}; thermodynamic defect is reported separately in Table~\ref{tab:thermo}.}
\label{tab:secondary}\centering\footnotesize\eqtable{table_secondary}{4}
\end{table}

\paragraph{Secondary-structure agreement.}
MORAD improves pairing MCC by 19.4\% over RIDE (from 0.6110 to 0.7295) and leads all four metrics: precision 0.7824, recall 0.6816, F1 0.7226, and MCC 0.7295, versus RIDE's 0.6472, 0.5717, 0.5939, and 0.6110 (Table~\ref{tab:secondary}). Neither EternaFold nor these structure-derived base pairs enter the reward, which scores pairing with RNAfold (Appendix~\ref{app:reward}); these pairing gains are therefore measured with tools outside the reward.

\begin{table}[b]
\caption{\textbf{Thermodynamic evaluation on 153 targets.} MFE and EFE are minimum and ensemble free energies (kcal/mol); NED is normalized ensemble defect; PE is positional entropy; MFEfreq is the probability of the minimum-energy structure \citep{lorenz2011vienna,zadeh2011ensemble}. MORAD is update 390. Bold marks the best mean; $\sim$ marks metrics without a preferred direction, shown for context. Appendix~\ref{app:thermoretest} reports all nine metrics and the native reference.}
\label{tab:thermo}\centering\footnotesize\eqtable{table_thermodynamics}{5}
\end{table}

\paragraph{Thermodynamic ensemble profile.}
Table~\ref{tab:thermo} reports the thermodynamic evaluation on all 153 targets. MORAD leads all three metrics with a preferred direction: NED drops from RIDE's 0.3906 to 0.2830, a 27.5\% reduction in normalized target-structure defect, PE from 0.3059 to 0.2102, and MFEfreq rises from 0.3350 to 0.4402, so the ensembles concentrate more tightly on the target \citep{mccaskill1990partition,lorenz2011vienna,zadeh2011ensemble}. Free energies are shown for context and are not ranked: MORAD's MFE/EFE, $-30.149/{-}30.937$ kcal/mol, are the most stable here yet the furthest from the native $-21.261$. Appendix~\ref{app:thermoretest} reports the remaining metrics: absolute ED, ensemble diversity, melting temperature, and GC content.

\subsection{Reward design}
Existing RL rewards for RNA design score tertiary structure alone; RIDER, for example, combines structural scores with threshold bonuses \citep{hu2026rider}. MORAD instead combines six bounded desirability scores, so that pairing agreement, ensemble behavior, and composition share the learning signal with tertiary geometry (Figure~\ref{fig:reward}). To isolate this choice, we post-train two policies from the same pretrained RIDE using the same data, optimizer, and schedule, and change only the reward: a structural-only reward of the kind used by existing work, and the MORAD reward. Table~\ref{tab:rewardcontrol} reports both policies alongside their pretrained starting point.

\begin{table}[t]
\caption{\textbf{Reward comparison under an identical training setup.} Two policies are post-trained from the same pretrained RIDE with the same data, optimizer, and schedule, and differ only in the reward; each is shown at the last checkpoint of its run. The structural-only reward matches the scope of existing RNA design RL rewards and is defined in Appendix~\ref{app:reward}. These results come from the controlled experiment's own evaluation run, so some entries differ from Tables~\ref{tab:tertiary}--\ref{tab:thermo}. Bold marks the best of the three for metrics with a preferred direction; MFE, GC fraction, and within-target Hamming diversity are shown for context.}
\label{tab:rewardcontrol}\centering\footnotesize\eqtable{table_reward_control}{3}
\end{table}

The MORAD reward leads on ten of the eleven metrics with a preferred direction: tertiary geometry (GDT-TS 0.3627 to 0.3694; TM-score 0.3194 to 0.3358; RMSD 8.369 to 8.207~\AA; C1$\prime$ TM 0.3818 to 0.3919), pairing agreement (MCC 0.6612 to 0.6978; F1 0.6568 to 0.6901), ensemble defect (ED/nt 0.3569 to 0.3256), recovery (0.5097 to 0.5124), and pairing entropy (0.2851 to 0.2800). MFE-structure frequency is the one measurement where the structural-only reward is higher, 0.3880 against 0.3847, and it comes with a shift in composition: that policy reaches GC fraction 0.5912 and MFE $-29.81$ kcal/mol, while the MORAD reward stays at 0.4970 and $-26.49$ kcal/mol, both closer to the native values of 0.5207 and $-21.26$ kcal/mol. Since only the reward differs, these differences follow from the reward design (Appendix~\ref{app:reward}).

\paragraph{Effect of the auxiliary terms.}
Although the auxiliary terms carry only 15\% of the nominal weight, the MORAD reward also improves the shared tertiary scores: geometric aggregation lowers the reward of candidates that align well but pair poorly.

\subsection{Held-out performance during training}
\begin{figure}[t]
\centering\includegraphics[width=0.80\linewidth]{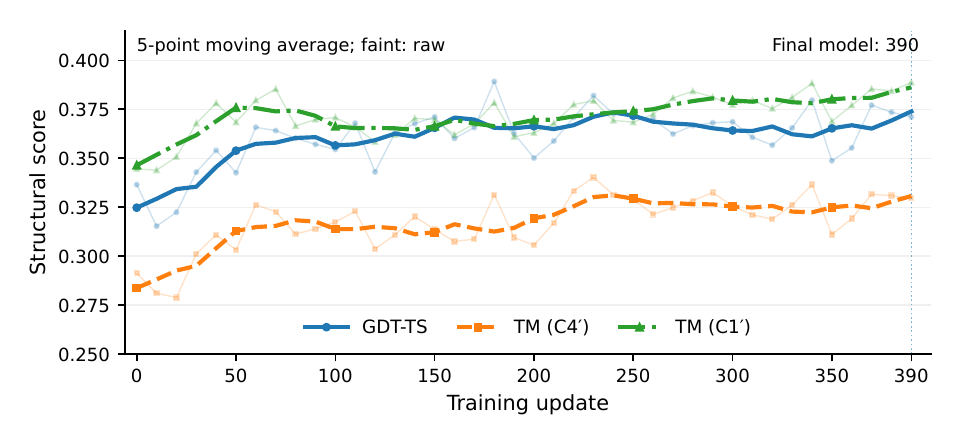}
\caption{\textbf{Structural-score trends over all 40 evaluations.} Thick lines show a centered five-point moving average; faint traces show the raw held-out scores recorded during training, with one sampled candidate per target. Near the two ends, the average uses only the available points. Table~\ref{tab:checkpoints} reports the evaluation of the last checkpoint, update 390. Appendix~\ref{app:trajectory} defines the smoothing.}
\label{fig:training}
\end{figure}

We evaluate the held-out targets every tenth update, giving 40 measurements from update 0 to 390 (Figure~\ref{fig:training}). GDT-TS increases from 0.3365 to 0.3711 and C4$\prime$ TM from 0.2912 to 0.3296: relative gains of 10.3\% and 13.2\%. These values come from the held-out evaluations run during training; Table~\ref{tab:tertiary} is a separate, final evaluation of update 390.

\paragraph{Candidate variation.}
During training, within-target sequence diversity decreases from 0.2023 to 0.1616. In the final eight-candidate evaluation, MORAD has Hamming and 3-mer diversity of 0.2344 and 0.3673, versus RIDE's 0.2971 and 0.5270; Appendix~\ref{app:seqdiv} gives all models. This quality--diversity tradeoff is well-documented in language-model RL and protein inverse folding \citep{kirk2024rlhfdiversity,yue2025rlvrboundary,ektefaie2024rldif,wang2025proteinzero}, motivating joint reporting of both dimensions.
\FloatBarrier

\section{Limitations}
The experiments evaluate MORAD on a single 10.2M-parameter RIDE backbone, demonstrating the framework on a compact model; the same modular interfaces accommodate larger structure-conditioned backbones. Reward scoring and tertiary evaluation both use the RhoFold+ predictor, consistent with current RNA design RL practice \citep{hu2026rider,shen2024rhofold}; experimental structural validation remains a separate step beyond the scope of this computational study. The reward also covers equilibrium folding properties, leaving kinetics and function-specific objectives as future additions to the reward.

\section{Conclusion}
MORAD establishes a shared-policy online post-training workflow for RNA design, integrating target-relative feedback, six-component rewards, and forward-process diffusion optimization. The last checkpoint of the completed run (update 390) reaches mean scores of 0.3827 GDT-TS and 0.3400 TM-score in the final evaluation. Together with the pairing and ensemble-defect results and the full training trajectory, these results document the quality gains and multi-objective tradeoffs of online post-training.
\label{main:end}
\subsection*{AI use statement}
Language-model assistance supported manuscript revision and translation, consistency checking, and code for tables, training plots, and document compilation. The two method diagrams (Figures~\ref{fig:method} and \ref{fig:reward}) were drawn by the authors. Empirical measurements come from the authors' experiment records; no new measurements or training runs were produced during manuscript preparation. The authors are responsible for the scientific claims, evidence, references, and final submission.

\subsection*{Ethics statement}
This is a computational study. It uses publicly available RNA structures from RNA3DB and the PDB and involves no human subjects, no animal experiments, and no new laboratory measurements \citep{szikszai2024rna3db,berman2000pdb}. Designed sequences are computational candidates scored by folding and thermodynamic software; none were synthesized or assayed, and we make no functional, therapeutic, or biosafety claims for them.

\subsection*{Reproducibility statement}
Algorithm~\ref{alg:train} and Eqs.~\eqref{eq:adv}--\eqref{eq:reward} specify the training procedure and the reward. Appendix~\ref{app:reward} lists every anchor and weight, Appendix~\ref{app:settings} the run configuration, and Appendix~\ref{app:protocol} the aggregation rules and valid-observation counts; Appendix~\ref{app:data} documents the data splits and screening. The complete training and evaluation code, the data, and the trained model, together with the trajectory, checkpoint, and measurement records and the table- and figure-generation scripts, are available from the project page.
\bibliographystyle{iclr2027_conference}
\bibliography{references}
\clearpage
\appendix
\section{Forward-process objective}
\label{app:derivation}
For one candidate and sampled noise level, write $b=\epsilon_{\bar\theta}-\epsilon$ and $\Delta=\epsilon_\theta-\epsilon_{\bar\theta}$. The two branches in Eq.~\eqref{eq:branches} give
\begin{align}
\ell_{\rm policy}
&=\frac{r\mse{b+\beta\Delta}+(1-r)\mse{b-\beta\Delta}}{\beta}\\
&=\frac{\mse{b}}{\beta}+2(2r-1)\langle b,\Delta\rangle_{\rm av}
 +\beta\mse{\Delta}.\label{eq:expansion}
\end{align}
Here the average divides by the number $D$ of scalar prediction coordinates. Treating the sampling-policy prediction $\epsilon_{\bar\theta}$ and the weight $r$ as constants,
\begin{equation}
\nabla_{\epsilon_\theta}\ell_{\rm policy}
=\frac{2}{D}\big[(2r-1)b+\beta\Delta\big].\label{eq:gradient}
\end{equation}
At $\Delta=0$, gradient descent moves the prediction toward the sampled noise for $r>1/2$ and in the opposite direction for $r<1/2$. At $r=1/2$, only the displacement penalty remains. The reference term in Eq.~\eqref{eq:loss} further penalizes deviation from the pretrained predictor. This expansion shows how the reward enters the DiffusionNFT-style update \citep{zheng2026diffusionnft}.

\paragraph{Parameterization and noise sampling.}
For a clean-sequence predictor $\widehat x_{0,\theta}$, the Gaussian parameterization \citep{ho2020ddpm} gives
\begin{equation}
\epsilon_\theta=\frac{z_t-\alpha_t\widehat x_{0,\theta}}{\sigma_t},
\qquad
\mse{\widehat x_{0,\theta}-x_0}
=\frac{\sigma_t^2}{\alpha_t^2}\mse{\epsilon_\theta-\epsilon}.
\end{equation}
The implementation uses uniform noise-space weighting with no residual normalization. Each candidate receives $K_t=8$ stratified log-SNR draws from $[-10,10]$. For log-SNR $\ell$, it uses $t=\operatorname{sigmoid}(-\ell)$, $\alpha_t=\sqrt{1-t}$, and $\sigma_t=\sqrt t$. Candidate one-hot sequences and feedback weights are treated as constants; gradients flow only through the current denoiser.

\section{Reward parameters and reference schemes}
\label{app:reward}
\begin{table}[htbp]
\caption{\textbf{Six-component reward parameters.} Anchors $a_j,b_j$ identify low and high desirability. Equation~\eqref{eq:desirability} uses floor $10^{-3}$. All three structural measurements must be finite; available auxiliary terms determine the active weights.}
\centering\small
\begin{tabular}{@{}l*{3}{c}@{}}
\toprule Measurement & $a_j$ & $b_j$ & $w_j$\\\midrule
GDT-TS & 0.20 & 0.75 & 0.25\\
TM-score & 0.20 & 0.70 & 0.30\\
RMSD (\AA) & 12.0 & 2.0 & 0.30\\
Secondary-structure MCC & 0.30 & 0.85 & 0.05\\
Ensemble defect / nucleotide & 0.60 & 0.15 & 0.05\\
Composition deviation & 0.50 & 0.02 & 0.05\\\bottomrule
\end{tabular}\label{tab:rewardanchors}
\end{table}

\paragraph{Auxiliary references.}
During training, RNAfold predicts both candidate and native secondary structures \citep{lorenz2011vienna}. Their base-pair sets define MCC \citep{matthews1975mcc}; the native RNAfold structure also defines the ensemble-defect reference \citep{zadeh2011ensemble}. For candidate and native base frequencies $p_b,q_b$, the implemented composition term is
\begin{equation}
C(y,y^\star)=\max\!\left(0,\sum_{b:p_b>0}p_b\log\frac{p_b}{\max(q_b,10^{-3})}\right).
\label{eq:composition}
\end{equation}
Logarithms are natural. An unavailable auxiliary measurement is removed from the active set, followed by the weight normalization in Eq.~\eqref{eq:reward}.

\paragraph{Structural-only control reward.}
The control scores tertiary geometry alone, matching the scope of existing RNA design RL rewards \citep{hu2026rider}: $R_{\rm ctrl}=0.45\phi_{.15,.70}(g)+0.35\phi_{.20,.70}(t)+0.20\phi_{12,1.5}(d)$, where $g,t,d$ are GDT-TS, TM-score, and RMSD, and $\phi_{a,b}(m)=\tfrac12[1-\cos(\pi\clip((m-a)/(b-a),0,1))]$. The MORAD reward of Eq.~\eqref{eq:reward} adds pairing agreement, ensemble defect, and composition on the same desirability scale. The two policies in Table~\ref{tab:rewardcontrol} differ in this reward alone; the Hamming diversity reported there averages pairwise normalized sequence distances within a target.

\paragraph{Published RIDER reward.}
\label{app:riderreward}
RIDER defines four structural base rewards $R_{\rm base}\in\{(5g)^2,(5t)^2,-(0.5d)^2,(5g)^2-(0.5d)^2\}$ with the published bonus \citep{hu2026rider}
\begin{equation}
R_{\rm RIDER}=R_{\rm base}+
\begin{cases}
100(g-0.5),&g>0.5,\\
20(2-d),&g\le0.5,\ d<2,\\
0,&\text{otherwise}.
\end{cases}\label{eq:riderreward}
\end{equation}
The released RIDER code uses the combined base reward with thresholds 0.45 and 3~\AA\ \citep{rider2026implementation}. This reward is defined and optimized for a single target backbone, whereas a shared policy requires a reward that is comparable across targets; the control above keeps the structural-only scope while meeting that requirement. Figure~\ref{fig:reward} compares the two designs.
\FloatBarrier

\section{Data construction and split characterization}
\label{app:data}

\paragraph{Records and screening.}
Each record contains a native sequence, its source PDB chain, backbone coordinates, and a coordinate-validity mask. Normalization maps letters to uppercase and T to U; scoring uses A/C/G/U. Residues with 3D coordinates are matched to sequence positions. Screening folds the native sequence with RhoFold+ and keeps targets whose prediction reaches C4$\prime$ GDT-TS of at least 0.30 under centered, index-preserving Kabsch alignment \citep{kabsch1976rotation}.

\paragraph{Fixed splits.}
The training set has 527 targets of length 27--258 nt (median 75); the evaluation set has 153 targets of length 15--186 nt (median 65). Evaluation sources comprise 71 RNA3DB entries, drawn from RNA3DB components that contain no training target, and 82 entries curated from RCSB \citep{szikszai2024rna3db,berman2000pdb}. All 153 are used for evaluation. When the native sequences are folded with RhoFold+, the median GDT-TS is 0.6400 for the training set and 0.6014 for the evaluation set.

\paragraph{Train--test separation.}
The two sets are built independently. Character-wise and MD5 checks report no exact match between the 153 normalized evaluation sequences and the 527 training sequences. To characterize the residual similarity that remains after exact-duplicate removal, we compute each evaluation target's maximum global identity to the training set under Needleman--Wunsch alignment \citep{needleman1970alignment}. The median is 0.6207 and the mean 0.6615; 26 of the 153 targets reach 0.80, including 12 that reach 0.95; at the pair level, however, only 64 of the 80,631 evaluation--training pairs (0.08\%) reach 0.80. All 153 targets are kept, with no exclusion based on similarity or results. Both policies in Table~\ref{tab:rewardcontrol} are trained on the same training set and evaluated on the same evaluation set, so the reward comparison holds this residual similarity fixed.

\paragraph{Reference use.}
Backbones and masks condition generation. Native RNAfold pairings and nucleotide frequencies provide training references. Pairing evaluation uses base pairs extracted from the native 3D structures, available for 137 targets; the other 16 short RNAs have no pairing reference and remain in the tertiary, recovery, and thermodynamic evaluations.
\FloatBarrier

\section{Implementation and evaluation protocol}
\label{app:settings}
\begin{table}[htbp]
\caption{\textbf{Settings of the reported run.}}
\centering\small
\begin{tabular}{@{}ll@{}}
\toprule Setting & Value\\\midrule
Base / parameters & Pretrained RIDE / 10,214,213\\
Training / evaluation targets & 527 / 153\\
Epochs / updates & 3 / 390\\
Targets / candidates per update & 4 targets $\times$ 8 candidates = 32\\
Learning rate / minimum ratio & $2\times10^{-4}$ / 0.3; 5\% linear warmup, then cosine\\
Hardware / synchronization & Four GPUs / distributed gradient averaging\\
Held-out evaluation & Every 10 updates, from 0 to 390; 40 observations\\
Evaluated checkpoints / reported & 180, 230, 340, 390 / 390 (last checkpoint)\\
Checkpoint selection & None: no validation split; the last checkpoint is reported\\\bottomrule
\end{tabular}
\label{tab:settings}
\end{table}
Shared settings are 50 sampling steps (each predicts $x_0$ and re-noises it) at sampling temperature 1, bounded to $[0.2,1.5]$ and raised by $0.1$ for the single resampling round, AdamW \citep{loshchilov2019adamw} with zero weight decay, gradient clipping at 1, $K_t=8$, $\beta=0.1$, and $\lambda_{\rm ref}=0.02$. All reported evaluations use the same RhoFold+ weights.

\subsection{Aggregation and valid observations}
\label{app:protocol}
For the main structural and pairing comparisons, candidate index 0 is taken for each target before aggregation. The index is fixed before scoring and is the same for every model, so each mean averages exactly one sample per target, with no selection. If $\mathcal T_m$ is the valid-target set for metric $m$, the reported mean is
\begin{equation}
\widehat\mu_m=\frac{1}{|\mathcal T_m|}\sum_{j\in\mathcal T_m}m_{j,0}.
\end{equation}
The evaluation set has 153 targets, with 126 valid C1$\prime$ TM measurements. No best-of-eight selection or within-target eight-candidate average is used in Tables~\ref{tab:tertiary} and \ref{tab:secondary}. The candidate-0 protocol yields one observation per target, so these aggregates are point estimates. Checkpoint means and standard errors (Appendix~\ref{app:checkpointdetails}) come from an earlier eight-candidate evaluation and are reported separately \citep{agarwal2021precipice}. The thermodynamic evaluation covers all 153 targets under its own protocol (Appendix~\ref{app:thermoretest}).

\subsubsection{Secondary-structure valid counts}
\label{app:paircounts}
\begin{table}[htbp]
\caption{\textbf{Valid target counts for the candidate-0 pairing metrics.} Counts correspond directly to Table~\ref{tab:secondary}; undefined measurements are excluded rather than replaced with zero. MORAD is update 390.}
\centering\small\eqtable{table_secondary_counts}{4}
\label{tab:paircounts}
\end{table}
Recall and F1 have 137 valid targets for every model. Precision and MCC use 136 for MORAD, 133 for RIDE, 132 for gRNAde ($T=0.1$), 135 for RDesign and 133 for RiboDiffusion. Precision and MCC are undefined for a design with no predicted base pair, whereas Recall and F1 remain defined; Recall and F1 therefore compare all models on the same 137 targets, and MORAD leads both (Table~\ref{tab:secondary}). For Precision and MCC, MORAD has the most valid targets.

\subsubsection{Sequence diversity across eight candidates}
\label{app:seqdiv}
\begin{table}[htbp]
\caption{\textbf{Sequence-diversity comparison.} Each model uses 153 targets and 1,224 candidates (eight per target). Bold marks the highest reported diversity; higher diversity alone does not establish greater structural accuracy.}
\centering\small\eqtable{table_sequence_diversity}{4}
\label{tab:seqdiv}
\end{table}
Hamming diversity summarizes within-target normalized sequence differences. MORAD has 0.2344/0.3673, intermediate between gRNAde ($T=0.1$)'s 0.1579/0.2406 and RIDE's 0.2971/0.5270. RDesign has the highest values, 0.4368/0.6184. These eight-candidate metrics are computed separately from the candidate-0 structural comparison and from the diversity curve recorded during training (Figure~\ref{fig:diversity}).
\FloatBarrier

\subsection{Metric definitions and interpretation}
\label{app:metrics}
\paragraph{Structure and pairing.}
C4$\prime$ RMSD is $\sqrt{L^{-1}\sum_i d_i^2}$ for aligned residue distances $d_i$. GDT-TS uses the fractions below 1, 2, 4, and 8~\AA\ \citep{zemla2003lga}; our implementation averages them under one index-preserving Kabsch alignment \citep{kabsch1976rotation}. The C4$\prime$ TM-score averages $[1+(d_i/d_0)^2]^{-1}$ after index-preserving Kabsch alignment, with the standard length-dependent $d_0$ \citep{zhang2004tmscore}. C1$\prime$ TM uses US-align \citep{zhang2022usalign}. Precision is $\mathrm{TP}/(\mathrm{TP}+\mathrm{FP})$, recall is $\mathrm{TP}/(\mathrm{TP}+\mathrm{FN})$, and F1 is $2\mathrm{TP}/(2\mathrm{TP}+\mathrm{FP}+\mathrm{FN})$; MCC applies the correlation coefficient to the base-pair confusion matrix \citep{matthews1975mcc}. EternaFold predictions cover canonical pairings \citep{waymentsteele2022eternafold}; the structure-derived reference base pairs include an estimated 12.8\% noncanonical pairs. The native sequences score F1/MCC of 0.6652/0.6713 under this evaluation.

\paragraph{Additional structural diagnostics.}
\label{app:structdiagnostics}
Table~\ref{tab:tertiary} also reports success rate, sequence recovery, C4$\prime$ lDDT, pLDDT and a coarse clash score. lDDT measures local distance agreement, whereas pLDDT is the predictor's confidence, reported on a 0--1 scale (e.g., 0.6726 for MORAD). The coarse clash score is a coarse-grained steric-clash measure from the evaluation pipeline, not an all-atom clashscore. The evaluation report does not state the success threshold, lDDT distance cutoffs, or clash cutoff, so these values are reported as given. RiboDiffusion leads pLDDT and recovery despite lower alignment scores, showing that these metrics measure different properties.

\paragraph{Omitted outputs.}
We omit three further outputs: Rfam covariance-model scores, because the model library was unavailable; target probability, because the implementation returned values above one; and INF-wc, because it duplicates MCC. No zeros or values from other models are substituted for them.

\paragraph{Target compatibility and ensemble concentration.}
Let $P_i(s)$ denote nucleotide $i$'s pairing-state probability, including the unpaired state, and $s_i^\star$ its target state. Ensemble defect and its normalized form are
\begin{equation}
\mathrm{ED}=\sum_{i=1}^L[1-P_i(s_i^\star)],\qquad
\mathrm{NED}=\mathrm{ED}/L.
\end{equation}
NED is dimensionless and supports comparisons across RNA lengths; ED counts expected incorrect nucleotides \citep{zadeh2011ensemble}. PE summarizes uncertainty in per-position pairing states; it is reported as computed by the evaluation pipeline, whose logarithm base was not recorded.

MFE is the energy of the minimum-energy secondary structure; EFE is the free energy of the Boltzmann ensemble. For one sequence and a consistent energy model, $\mathrm{MFEfreq}=\exp[-(\mathrm{MFE}-\mathrm{EFE})/(RT)]$ \citep{mccaskill1990partition,lorenz2011vienna}. The table reports each measured mean directly rather than deriving a mean frequency from mean energies. Lower PE and higher MFEfreq indicate a more concentrated ensemble; NED directly assesses target compatibility. Lower modeled energy alone does not establish target accuracy or folding kinetics.

\subsection{Complete thermodynamic evaluation}
\label{app:thermoretest}
\begin{table}[htbp]
\caption{\textbf{Complete thermodynamic measurements on 153 targets.} MFE/EFE: kcal/mol; ED: nt; NED: dimensionless; EDiv: mean base-pair distance; PE: positional entropy; MFEfreq: probability; $T_m$: $^\circ$C; GC: fraction. MORAD is update 390. Bold marks the best of the six models in the indicated direction; Native is a reference and is not ranked. Dashes denote unreported reference values; $\sim$ marks metrics without a preferred direction, shown for context.}
\centering\small\setlength{\tabcolsep}{3pt}
\eqtable{table_thermodynamics_full}{7}
\label{tab:thermofull}
\end{table}
EDiv measures the mean distance between secondary structures within the Boltzmann ensemble, not diversity among generated sequences \citep{lorenz2011vienna}. Larger values mean the ensemble contains more varied structures, which need not improve target accuracy, so EDiv is shown for context and not ranked. MORAD has EDiv 5.605 versus RIDE's 7.512, consistent with a more concentrated ensemble. $T_m$ (predicted melting temperature) and GC content are reported as descriptors: MORAD has 72.52~$^\circ$C and GC 0.5393, versus 72.54~$^\circ$C and 0.5400 for RIDE. The native reference has ED 28.049 and NED 0.3741; it is a sequence reference rather than a guaranteed optimum under the evaluation model.
\FloatBarrier

\section{Checkpoint comparison and additional results}
\label{app:checkpointdetails}
\begin{figure}[!b]
\centering\includegraphics[width=.66\linewidth]{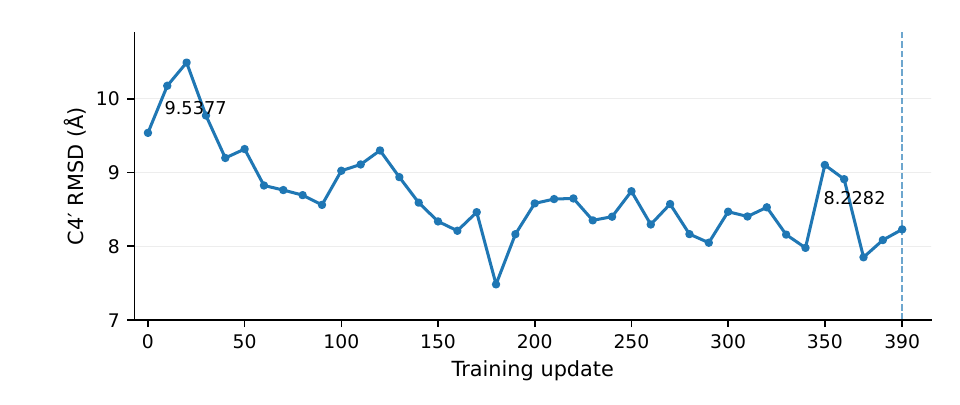}
\caption{\textbf{Complete RMSD trajectory.} All 40 held-out measurements are connected in time order. Lower RMSD indicates closer agreement after alignment; the vertical marker shows update 390.}
\label{fig:rmsd}
\vspace{8pt}
\centering\includegraphics[width=.66\linewidth]{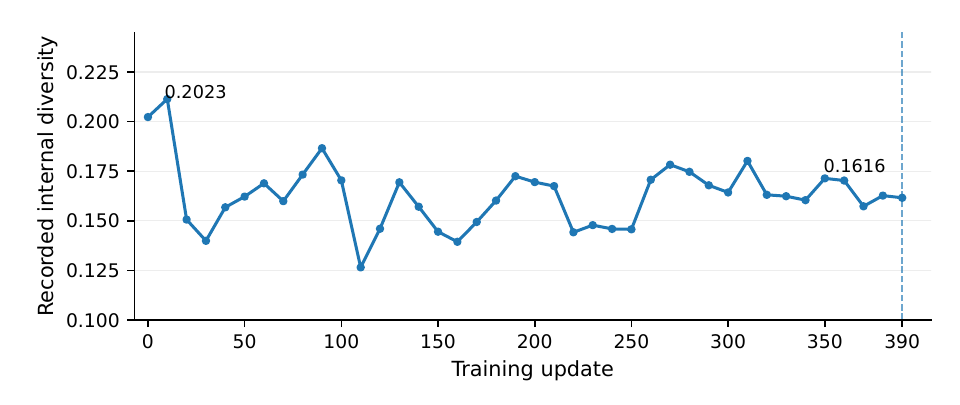}
\caption{\textbf{Within-target diversity during training.} Diversity decreases from 0.2023 to 0.1616. The eight-candidate comparison across models is in Table~\ref{tab:seqdiv}.}
\label{fig:diversity}
\end{figure}
\begin{table}[htbp]
\caption{\textbf{Checkpoint comparison.} Means over eight candidates per target, with standard errors, for four checkpoints from an earlier evaluation run; they differ from the candidate-0 results in Tables~\ref{tab:tertiary} and \ref{tab:secondary}, which also report update 390, the last checkpoint.}
\centering\small\eqtable{table_checkpoints}{5}
\label{tab:checkpoints}
\end{table}
Update 390 is the last checkpoint and leads all five means in this evaluation. Its eight-candidate GDT/TM values (0.3774/0.3338) are kept here for reference; the main comparison uses the candidate-0 values (0.3827/0.3400).

\subsection{Training diagnostics}
\label{app:trajectory}
Figures~\ref{fig:rmsd} and \ref{fig:diversity} show all 40 evaluations run during training. Over these evaluations, recovery changes from 0.5020 to 0.5069 and success rate from 0.2288 to 0.3137.

\paragraph{Smoothing.}
For a raw structural score $q_j$ at update $10j$, $j=0,\ldots,39$, Figure~\ref{fig:training} displays
\begin{equation}
\widetilde q_j=\frac{1}{|W_j|}\sum_{k\in W_j}q_k,\qquad
W_j=\{\max(0,j-2),\ldots,\min(39,j+2)\}.
\end{equation}
All three structural scores use this window. The raw values remain visible in the figure; the diagnostics in this appendix are not smoothed.

\FloatBarrier

\end{document}